\documentclass[conference]{IEEEtran}
\IEEEoverridecommandlockouts

\usepackage{cite}
\usepackage{amsmath,amssymb,amsfonts}
\usepackage{algorithmic}
\usepackage{graphicx}
\usepackage{textcomp}
\usepackage{xcolor}
\usepackage[flushleft]{threeparttable}
\usepackage[normalem]{ulem}

\usepackage{amsmath,amssymb,amsfonts}
\usepackage{algorithmic}
\usepackage{multirow}

\usepackage{threeparttable}
\usepackage{tabularx}
\usepackage{siunitx}
\usepackage{amssymb}
\usepackage{pifont}

\usepackage{hyperref}
\usepackage{cleveref}
\usepackage{bm}

\def\BibTeX{{\rm B\kern-.05em{\sc i\kern-.025em b}\kern-.08em
    T\kern-.1667em\lower.7ex\hbox{E}\kern-.125emX}}

\usepackage{fancyhdr}   
                            
\fancypagestyle{firstpage}{     
  \fancyhf{}
  \chead{\textcolor{gray}{This article has been accepted for publication in the proceedings of the \\ 2023 IEEE International Workshop on Advances in Sensors and Interfaces (IWASI).}}
  \fancyfoot[C]{\small{\textcolor{gray}{~\copyright~ 2023 IEEE.  Personal use of this material is permitted.  Permission from IEEE must be obtained for all other uses, in any current or future media, including reprinting/republishing this material for advertising or promotional purposes, creating new collective works, for resale or redistribution to servers or lists, or reuse of any copyrighted component of this work in other works.}}}

}

\begin{document}


\title{ColibriUAV: An Ultra-Fast, Energy-Efficient Neuromorphic Edge Processing UAV-Platform with Event-Based and Frame-Based Cameras \\
\thanks{This work was supported by the Innosuisse project Eye-Tracking 103.364 IP-ICT with Agreement Number: 2155011780, and the Innovation Programme APROVIS3D under the Grant 20CH21\_186991.}

}

\author{\IEEEauthorblockN{ Sizhen Bian\IEEEauthorrefmark{1},  Lukas Schulthess\IEEEauthorrefmark{1}, Georg Rutishauser\IEEEauthorrefmark{1}, Alfio Di Mauro\IEEEauthorrefmark{1}, Luca Benini\IEEEauthorrefmark{1}\IEEEauthorrefmark{2}, Michele Magno\IEEEauthorrefmark{1}}
\IEEEauthorblockA{\IEEEauthorrefmark{1}\textit{Departement Informationstechnologie und Elektrotechnik, ETH
    Z{\"u}rich, Switzerland}}
\IEEEauthorblockA{\IEEEauthorrefmark{2}\textit{Dipartimento di Ingegneria dell'Energia Elettrica e dell'Informazione, Universit\`{a} di Bologna, Bologna, Italy}}
}



\maketitle


\begin{abstract}
The interest in dynamic vision sensor (DVS)-powered unmanned aerial vehicles (UAV) is raising, especially due to the microsecond-level reaction time of the bio-inspired event sensor, which increases robustness and reduces latency of the perception tasks compared to a RGB camera. 
This work presents ColibriUAV, a UAV platform with both frame-based  and event-based cameras interfaces for efficient perception and near-sensor processing. The proposed platform is designed around Kraken, a novel low-power RISC-V System on Chip with two hardware accelerators targeting spiking neural networks and deep ternary neural networks.Kraken is capable of efficiently processing both event data from a DVS camera and frame data from an RGB camera. A key feature of Kraken is its integrated, dedicated interface with a DVS camera. This paper benchmarks the end-to-end latency and power efficiency of the neuromorphic and event-based UAV subsystem, demonstrating state-of-the-art event data with a throughput of 7200 frames of events per second and a power consumption of 10.7 \si{\milli\watt}, which is over 6.6 times faster and a hundred times less power-consuming than the widely-used data reading approach through the USB interface. The overall sensing and processing power consumption is below 50 mW, achieving latency in the milliseconds range, making the platform suitable for low-latency autonomous nano-drones as well.

\end{abstract}

\begin{IEEEkeywords}
Event camera, event interface, autonomous drone, on-the-edge computing, autonomous navigation 
\end{IEEEkeywords}

\section{Introduction}

\thispagestyle{firstpage}

The autonomous navigation of agile UAVs requires fast response times from environment perception to motor instructions, as well as high energy efficiency to handle complex computing tasks. In recent years, various sensing modalities have been explored for reliable autonomous decision-making, including UWB~\cite{niculescu2022energy}, Time-of-Flight (ToF) sensors~\cite{muller2022fully}, and radar~\cite{miccinesi2022geo}. However, the most popular sensing approach for local positioning and obstacle avoidance of drones is still vision sensors such as RGB cameras, which provide richer contextual information and unlock more functionalities of UAVs than other sensing modalities \cite{muller2022fully}.

Despite the advantages of vision sensors, the volume of streamed vision data is large and challenges the memory footprint of limited hardware resources of a miniaturized UAV during data processing. Researchers have thus struggled to compress image-processing models to fit onboard for fast and energy-efficient data processing \cite{suleiman2019navion, lamberti2022tiny}. 

Event-based cameras, also called dynamic vision sensors, show promise for improving energy efficiency and latency in robotic applications \cite{gallego2020event}. By exploiting the bio-inspired principle of sparse event train where an event is only generated when a pixel's brightness changes, the event camera generates much less redundant data, which is attractive for a resource-limited edge platform. The microsecond level of temporal resolution and sparsity of the output enables faster motion capturing without introducing motion blur~\cite{haoyu2020learning}. Moreover, pixels have a logarithmic response to the brightness signal, which allows the event camera a very high dynamic range, being able to see dark and bright regions simultaneously~\cite{rebecq2019high}. The output of a DVS is based on a threshold mechanism: each pixel will memorize the logarithm of the brightness when an event is triggered and continuously monitor the brightness change respective to the memorized value. When the difference crosses the thresholds, the next event is triggered, and the stored brightness value is updated~\cite{delbruck2020v2e}. The transmitted events consist of the pixel's location, the timestamp, and the polarity of the change, where the increase of brightness is represented with an \textit{ON} event and the decrease with an \textit{OFF} event.

Today’s event-based cameras are still characterized by high power consumption due to their IO interface, in contrast to the low power consumption of the sensors~\cite{di2021flydvs}. This is especially due to the non-standard communication protocol they adopt. Today's DVS cameras connect with external chips through a USB interface, reducing the benefits of energy efficiency and latency in real end-to-end application scenarios ~\cite{falanga2020dynamic, mueggler2015towards}.

However, USB DVS cameras are already used to collect datasets and in UAV applications, often coupled with GPUs \cite{stoffregen2019event}. Nevertheless, considering the particularity of the event data format in the spatial and temporal domains, to fully leverage the benefits of the bio-inspired event-based paradigm, new models and algorithms are needed to extract typical features from the asynchronous and high-temporal resolution event streams in the spatial and temporal domains. Additionally, efficient neuromorphic platforms are required to handle the event streams from the DVS and enable the realization of the paradigm's performance potential\cite{gruel2022neuromorphic}.

\begin{table*}[!t]
      \caption{Drones equipped with dynamic vision sensors (DVS)}
      \label{tbl:drone_dvs_papers}
    \begin{threeparttable}
      \begin{tabular}{ p{1.2cm} p{1.5cm} p{1.2cm} p{1.2cm} p{1.5cm}p{1.7cm} p{1.2cm} p{1.2cm} p{1.2cm} p{1.7cm}}
      \hline
        & Computing Unit & Event Camera & interface & Algorithm  & $P_{inf}$ (\si{\milli\watt}) \tnote{a} & $T_{inf}$ (\si{\milli\second}) \tnote{a}  &  $T_{samp}$ (\si{\milli\second}) \tnote{b} &  $T_{close\_loop}$ (\si{\milli\second}) \tnote{c} &  $P_{close\_loop}$ (\si{\milli\watt}) \tnote{c}
        \\
        \hline 
        \cite{falanga2020dynamic}-2020 & NVIDIA Jetson TX2 & two SEES1 & USB2.0 & Ego-motion compensation   & 1000+ (Watt level) & 3.56 & 10 & N/A \tnote{d} & 1000+ (Watt level) \\
        \hline
        \cite{andersen2022event}-2022 &  NVIDIA Jetson TX2 & DAVIS2404 & USB2.0 & Sparse Gated Recurrent Network & 1000+ (Watt level) & 500 & $\approx$ 1000- 2000 & N/A  & 1000+ (Watt level) \\
        \hline
        \cite{mueggler2015towards}-2015 &  Odroid U3 quad-core computer & two DVS128 & USB2.0 & event-based circle tracker, EKF &  1000+ (Watt level)  & 4.5 & N/A & N/A  & 1000+ (Watt level) \\
        \hline
        \cite{dimitrova2020towards}-2020  & UP Board & DAVIS 240C & USB2.0 & hough transform, kalman filter &  1000+ (Watt level)  & 12 & 3 & N/A  &  1000+ (Watt level) \\
        \hline
        \cite{vitale2021event}-2021 & Loihi, KapohoBay & DAVIS 240C & SAER \tnote{e} & Hough transform   &  1000+ (Watt level)  & 0.25 & 0.15 & N/A  &  1000+ (Watt level)  \\
        \hline
        \cite{chen2022esvio}-2022 & Intel NUC computer & two DAVIS346 & USB3.0 & graph-based optimization  &  1000+ (Watt level) &  N/A & N/A & N/A &  1000+ (Watt level)  \\
        \hline
        \textbf{Ours} & \textbf{Kraken} & \textbf{DAV132S} & \textbf{SAER} & \textbf{SNN} & \textbf{35.6}\tnote{f} & \textbf{163 (131+32)}\tnote{f} & \textbf{300} & \textbf{163 (131+32)}\tnote{h} & \textbf{46.98 (35.6 + 11.076 + 0.3)}\tnote{f} \\
        \hline
        \end{tabular}

      \begin{tablenotes}
            \item[a] $P_{inf}$, $T_{inf}$: Processing unit power/latency during inference of each instance. The power includes both idle state power and active state power.
            \item[b] $T_{samp}$: The window of each instance, includes both data collection and preprocessing.
            \item[c] Latency and power consumption from event perceiving to motor command giving.
          \item[d] Not available. 
          \item[e] Synchronous address event representation interface.
          \item[f] Inference time and power including the preprocessing on the fabric controller, close loop time and power including the DVS camera and the interfaces. See Table \ref{tbl:benchmarkingColibriUAV} for details.
        \end{tablenotes}  
      \end{threeparttable}
\end{table*}

One approach to processing event data is the spiking neural network (SNN)\cite{ponulak2011introduction, lee2016training}, which closely mimics the natural neural network, transmitting information only when the membrane potential of the neuron reaches the specific threshold. Then the neuron fires, generating a spike that is propagated to other neurons. SNNs preserve the temporal context in the spike train and attempt to exploit the salient information from spike numbers or rates in a certain window. Because of their non-differentiability, the backpropagation mechanism cannot be directly applied to training SNNs. Thus, surrogate gradients were explored for learning SNNs' synaptic weight and axonal delay parameters, like SLAYER \cite{shrestha2018slayer} or STBP\cite{wu2018spatio}. These approaches take much time for training but make full use of the temporal resolution of the data, achieving promising results in accuracy and latency~\cite{tan2021improved, taunyazov2020fast}.


On the hardware side, over the years, different computing platforms have been used for interpreting the event data input, including traditional digital processors such as Nvidia Jetson~\cite{falanga2020dynamic} and a new generation of neuromorphic processors, designed in both academia and industry, and examples are TrueNorth~\cite{akopyan2015truenorth} from IBM, Loihi~\cite{davies2018loihi} and its platform Kapoo bay \cite{vitale2021event} from Intel, and DYNAP-SE~\cite{moradi2017scalable} from ETH Zürich. Neuromorphic platforms are designed to use artificial neurons to do computations and have the potential to push the envelope of energy efficiency and execution speed \cite{davies2018loihi}. However, these neuromorphic chips support general SNN processing and are not necessarily optimized for edge event data processing in a seamless neuromorphic path from event perception to decision-making.



In 2022, researchers from ETH Zürich published an embedded heterogeneous system on chip (SoC) RISC-V based, called Kraken \cite{di2022kraken}, which integrates a neuromorphic accelerator, sparse neural engine (SNE) \cite{di2022sne}, aiming to accelerate SNN inference tasks at the extreme edge efficiently. Kraken is designed to optimize energy efficiency and low latency, and it embeds two on-chip camera interfaces: one for shutter-based cameras and another for event-based cameras, linking a ternary deep neural network on frame-form data and a spiking neural network for event-form data. The Kraken platform enables seamless neuromorphic processing of event data from perception to decision-making at the extreme edge, and it represents a significant step towards the practical implementation of end-to-end energy-efficient, bio-inspired event-based vision systems.


This work presents the design and development of ColibriUAV, which is the first energy-efficient drone platform based on the Kraken SoC. The platform includes both an event-based camera and an RGB camera to evaluate end-to-end latency, power, and energy efficiency. The main goal of the paper is to provide an experimental platform that exploits an event-based camera and neuromorphic accelerator for future UAV-related tasks and eventually create agile UAV systems with increased robustness under aggressive and jerky maneuvers.

The main contributions of this paper are as follows:

\begin{itemize}
    \item  The design and implementation of ColibriUAV, the first neuromorphic deep learning edge platform in the form of a drone. It includes a low-power event camera and a low-power heterogeneous SoC that integrates a sparse neuron engine for event train processing. The platform also preserves the frame camera interface for future evaluation.
\item Profiling and evaluating the latency and energy of the event-based subsystem. With experimental results, the paper shows that ColibriUAV achieves the state-of-art event data reading interface with 7200 event-frames per second throughput and 10.7 milliwatt power consumption, which is over 6.6 times faster and a hundred times less power than the widely used data reading approach through a USB interface.

\item Benchmarking the close-loop neuromorphic performance of ColibriUAV using a reference neural network featuring a latency of 163 ms, power of 46.98 mW, and energy consumption of 9.224 mJ for the end-to-end task from events to motor control. 

\end{itemize}

\section{Related work}
In recent years, research on low-latency event camera-embedded drones has rapidly grown, as summarized in \Cref{tbl:drone_dvs_papers}.
Falanga et al. \cite{falanga2020dynamic} utilized two SEES1 event cameras on a quadrotor and leveraged the temporal information contained in the event stream to distinguish between static and dynamic objects, with the aim of avoiding approaching obstacles with fast response times. They reported an overall latency of only 3.5 milliseconds, which is sufficient for reliable detection and avoidance of fast-moving obstacles.
Andersen et al. \cite{andersen2022event} proposed sparse convolutions for detecting gates in a race track using event-based vision, and achieved success rates between 0.2 to 1.0 for gate detection with different angular rates and scene illumination levels.
Mueggler et al. \cite{mueggler2015towards} proposed a method that tracks spherical objects on the image plane using probabilistic trackers that are updated with each incoming event. They experimentally demonstrated that the method enables initiating evasive maneuvers early enough to avoid collisions.
Dimitrova et al. \cite{dimitrova2020towards} explored one-dimensional attitude tracking using a dualcopter platform equipped with an event camera and reported promising results of the event-camera-driven closed-loop control. The state estimator performs with an update rate of 1 kHz and a latency of 12 milliseconds.
The first work of a neuromorphic vision-based controller on a chip, solving a high-speed UAV control task, is shown in \cite{vitale2021event}, where the DVS events are used as inputs to the neuronal cores of a neuromorphic chip using an address event representation interface and processed directly by the SNN. The proposed visual processing, from incoming events to estimated angle, on Loihi takes only 0.25 milliseconds.
The first event-based stereo visual-inertial odometry is described in \cite{chen2022esvio}, which leverages the complementary advantages of event streams, standard images, and inertial measurements for robust state estimation. The method was successfully validated on a quadrotor platform under low-light environments.

Although different focuses have been put into the combination of drones and event cameras, none of the existing works emphasize the performance increase in latency and power that neuromorphic form sensing and computing can bring. They mainly deployed the algorithms on more general platforms like GPUs or small-form computers, which heavily increase the load of a drone and limit its agility. For example, KapohoBay, used in \cite{vitale2021event}, shows impressive inference speed benefiting from the onboard spike processing. However, the KapohoBay platform consumes hundreds of milliwatts of power even in idle state, thus not suitable for practical deployment on drones.

This paper presents the design and development of a milliwatt drone that features a System on Chip (SoC) with dual camera interface and two hardware accelerators. In addition, it includes an 8-core RISC-V parallel processor for other tasks and embedded control.


\begin{figure}[!t]
\begin{minipage}[t]{1.0\linewidth}
\centering
\includegraphics[width=0.95\textwidth]{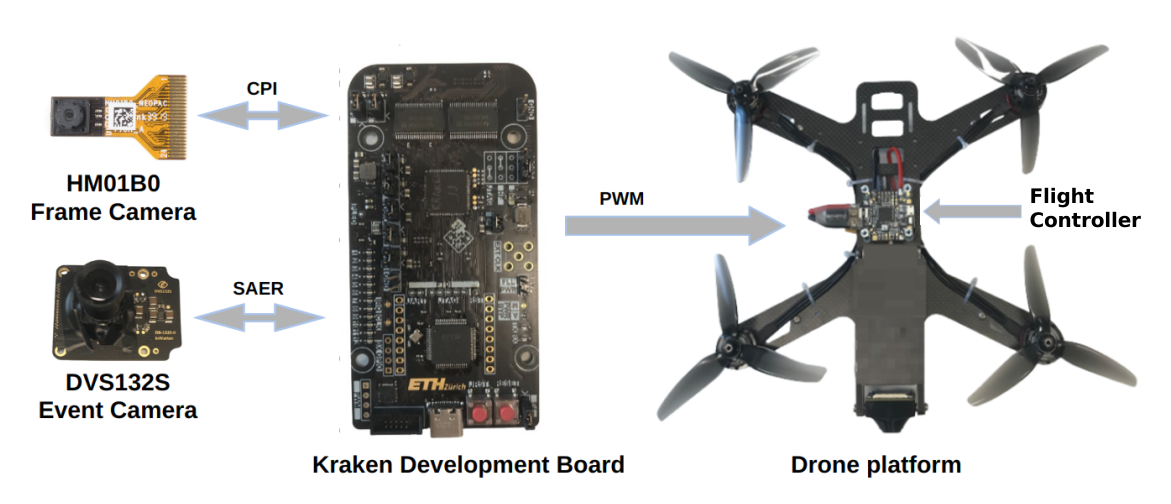}
\caption{System architecture: ColibriUAV with Dual Camera and Kraken Development Board.}
\label{structure}
\end{minipage}
\end{figure}

\begin{figure}[!t]
\begin{minipage}[t]{1.0\linewidth}
\centering
\includegraphics[width=0.85\textwidth]{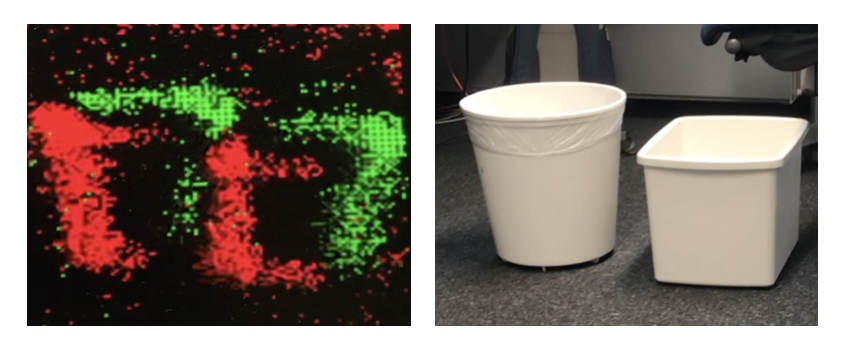}
\caption{Recordings at one particular event frame of the DVS132S (left) and an RGB camera (right): a scene with two bins.}
\label{Event_Frame_Barrel}
\end{minipage}
\end{figure}

\section{System Architecture}

In this section we detail the design and implementation of ColibriUAV, an end-to-end edge vision UAV system that includes both perception, computing, and actuation, as shown in Figure \ref{structure}. The platform can connect both the event camera and the frame camera directly to the Kraken SOC through an interface. In this section, we describe the core components used for the proposed ColibriUAV. The platform is built upon the commercially available drone frame Outlaw 270 racer, which uses four powerful motors of type RSII 2306 2400KV from Emax with a maximum thrust of \SI{1720}{\gram}.

The onboard flight controller F722-SE from Matek, together with the brushless electronic speed controller Xrotor Micro 40A from Hobbywing, implements the basic control functionalities. A 4S LiPo battery pack of type Nano-tech with a capacity of \SI{4500}{\milli\ampere\hour} provides the energy for autonomous operation.

Although ColibriUAV can host both an RGB camera and an event-based camera, this paper focuses on the event-based camera to evaluate its latency and energy-efficiency.


\section{Design and Implementation}

\subsection{ColibriUAV Platform}

ColibriUAV is designed around Kraken to exploit the data from event and frame cameras, process the information, and supply controlling information to the UAV pilot. It is designed to fit on the drone's frame while providing full flexibility and adaptability for even smaller nano drones.
The SAER interface is supported by an 80-pin board-to-board connector for data transfer with the DVS132S camera. Besides this, a CPI camera connector is also available for frame-based vision data receiving. Power of different domains of the Kraken chip, including the always-on fabric controller, power-switchable cluster, and independent power-switchable accelerators (SNE and CUTIE), is supplied by individual buck converters.

\subsection{Event-based Camera}

For the design, we selected the event camera DVS132S from iniVation AG, for its low power and real-time high-speed \cite{li2019132}.
It features a resolution of 132x104 with \SI{10}{\micro\meter} pixel side length on a \SI{65}{\nano\meter} process and a synchronous address-event representation (SAER) readout capable of 180 Meps (million events per second) throughput.
The power efficiency, compared to prior versions~\cite{brandli2014240, son20174}, has been mainly achieved by changing to a lower supply voltage of \SI{1.2}{\volt}.
For example, the chip consumes only \SI{250}{\micro\watt} with 100 Keps running at 1 K event frames per second.
With on-chip digital circuitry, the whole pixel array can be read synchronously, producing event frames when a \textit{SAMPLE} request occurs. The event frame is transferred into the pixel memory and then streamed out following the X\textbackslash Y scan clock. The event frame rate is thus dependent on the \textit{SAMPLE} request frequency. 

The \textit{ON/OFF} event of each pixel is generated depending on whether the changed luminosity of this pixel is larger than the pre-configured threshold.
Both the number of possible events in each frame (from 0 to 13728) and the sample rate influence the power consumption of the chip. The events are streamed through the SAER port, which also supports pre-readout pixel-parallel noise and spatial redundancy suppression, aiming to eliminate the redundant events caused mainly by flicker. It is worth noting that there are two simultaneous streams of output data: the X/Y address with one byte and the ON/OFF events with one byte recording the events of four pixels (2x2). Thus, to stream out all events, 13728 events in one frame, only 3432 (66x52) system clocks for one frame are needed (when not considering the sparse auxiliary clocks).
Figure \ref{Event_Frame_Barrel} shows an event-frame reading of the DVS132S at a single time step while in motion. The red dots represent the \textit{ON} events, while the green dots represent the \textit{OFF} events.

\subsection{Kraken SoC}


The Kraken is the core of ColibriUAV. Kraken is a heterogeneous SoC \cite{di2022kraken} optimized for power-efficient edge computing, particularly for event- and frame-based low-power visual processing by integrating acceleration engines. Kraken is composed of three subsystems: one fabric controller (a 32bit RI5CY/CV32E40P core), a parallel ultra-low-power (PULP) cluster with eight 32bit RISC-V cores, and a dedicated accelerator domain that hosts two dedicated hardware accelerators: SNE (Sparse Neural Engine) and CUTIE (Completely Unrolled Ternary Inference Engine).

The fabric controller acts as the main programmable control unit, hosting the main interconnection busses towards the main L2 memory and the advanced peripheral bus (APB), which controls all SoC peripherals. The computing cluster implements dedicated Instruction-Set Architecture (ISA) extensions like hardware loops, multiply and accumulate, and vectorial instructions for low-precision machine learning workloads. A 128 KiB tightly coupled data memory (L1) is shared among the cores to serve all memory requests.

The dedicated accelerator domain is composed of SNE targeting spiking convolutional neural network (SCNN) inference, and CUTIE, a TNN (ternary) accelerator designed to maximize energy efficiency by minimizing data movement during inference. To be noticed, the accelerator domain has two clock sources: one is from the fabric controller, which is used to clock the interface logic between the accelerators and the system interconnect, and the other is a dedicated one to clock the accelerators' computing engines. 




\section{Experimental Results}

An edge vision system's power consumption and closed-loop latency are mainly composed of four parts: the sensor, the processor, the actuator, and the interface between them for data/command transmission. For instance, a sensor with smaller resolution and lower output precision, and a shorter data/command transmission distance, results in less power consumption and faster temporal response. In this section, we evaluate the power consumption and latency separately for the DVS camera, Kraken, and interfaces in the proposed ColibriUAV. As the power and latency of the actuator, i.e., the motor and flight controller, highly depend on different UAV platforms targeting different application scenarios, this work does not evaluate this part.

\subsection{DVS Camera}
The DVS132S is a variant of the original iniVation DVS, which has an analog pixel with per-pixel timestamping at the output and is optimized for very low power consumption. We measure the power consumed by the DVS132S sensing module by measuring the voltage of two low-value resistors located before the digital and analog voltage inputs to the sensor, separately. The analog input shows a constant power consumption of 0.36 \si{\milli\watt} (which can vary depending on the bias configuration), while the digital input shows ultra-small power consumption up to 0.06 \si{\milli\watt} depending on the number of events, which is much less than the power consumption of other DVS cameras~\cite{dvsserie}.

As the event-frame sampling rate is configured on the Kraken chip and influences the power consumed by the DVS132S, the sampling rate was set to \SI{7.2}{\kilo\hertz} while the Kraken SoC runs at \SI{50}{\mega\hertz} system clock. The time spent to receive one event frame from the camera to the fabric controller of Kraken is thus a maximum of only \SI{139}{\micro\second} (one full-events frame), thanks to the SAER interface that groups events from all pixels in a frame. This is one of the main differences between our UAV platform and other UAVs incorporating DVS, which usually use the USB port to transfer data from the camera to the processing unit. 

The time spent on USB depends on the number of events in each event frame. USB 2.0 is specified for 480 Mbit/s. An event from the DVS is encoded in 32 bits, yielding a transfer duration of 0.067 µs \cite{mueggler2015towards}. Therefore, each event frame will take a maximum of 920 µs. In \cite{di2021flydvs}, the authors built an event data streaming node with a low-power FPGA that deployed a DVS driver module that physically interacts with the SAER interface of the DVS132S camera. The low-power FPGA can reach a reading speed of 874 Efps consuming only \SI{17.6}{\milli\watt} of power. \Cref{tbl:interface} summarizes the throughput and power of different interface solutions. Our platform achieves very high throughput with nearly one million events per second, which would take more than 6.6 seconds by USB and 8.2 seconds by SAÉR accomplished on low-power FPGA \cite{di2021flydvs}. The power consumption for event data fetching also shows impressive efficiency, consuming only \SI{10.7}{\milli\watt}, while USB-based solutions consume power on the order of watts \cite{berner20075}.

\begin{table}[!t]
\caption{DVS interface}
\label{tbl:interface}
\centering
\begin{threeparttable}
\begin{tabularx}{\linewidth}{{ p{3.0cm} p{3.0cm} p{1.7cm} }}
\hline
Modules\tnote{a}  & Throughput\tnote{a} (efps\tnote{b}) & Power\tnote{c}(\si{\milli\watt}) \\\hline
USB \cite{andersen2022event, mueggler2015towards, dimitrova2020towards, chen2022esvio} & 1087 &  over 1000 \tnote{d} \\ \hline
SAER on FPGA \cite{di2021flydvs} & 874 &   17.6 \\ \hline
\textbf{SAER on ColibriUAV }  & \textbf{7200}  & \textbf{10.656\tnote{e}}\\\hline

\end{tabularx}
\begin{tablenotes}
    \item[a] Assuming fully populated event frame (13728 events). 
    \item[b] Event-frames per second. 
    \item[c] Power consumption of the host platform when fetching data from the DVS camera (power consumption of DVS camera is not considered). 
    \item[d] In the level of \si{\watt}, depending on the host platform like Loihi KapohoBay, Jetson TX2, Odroid computer, etc. 
    \item[e] Measured with \SI{0.9}{\volt} for FC and \SI{1.8}{\volt} for IO domain on Kraken, the system clock is \SI{50}{\mega\hertz}. 
\end{tablenotes}
\end{threeparttable}
\end{table}

For the performance evaluation of the neuromorphic algorithm on the SNE on Kraken, a proof of concept network has been evaluated in our previous work~\cite{rutishauser2023colibries}. The spiking neural network is composed of two convolutional layers and two fully-connected layers and trained with the spatiotemporal backpropagation (STBP) described in~\cite{wu2018spatio} in a quantization-aware way, and then deployed in a layer-by-layer fashion. This network does not have the goal of evaluating the accuracy for a specific application task (such as object avoidance), but rather benchmarking the latency and energy for a medium-sized network. To configure the SNE to execute a specific layer of an SNN, the first step is to configure the engine and a few hyperparameters that are needed for each layer, including the base potential at which neurons are reset after firing, the firing threshold of the neurons, the rate at which the threshold is adapted, the minimum time between two consecutive spikes of the same neuron, and the amount of shifting applied to each timestep. Then, the trained weights are converted to SNE’s internal format and loaded into the engine's kernel memories. Finally, the streamers are configured to trigger the execution of SNE over a stream of events. The input streamer is configured to start loading the input events to the accelerator, and the output streamer is configured to specify the last expected event and a pointer to an accessible memory region that will be filled by produced events. During the layer execution, a tiling method is used to adjust the layer processing to the limited resources available on SNE.

\subsection{Motor control with PWM and end-to-end evaluation}
For the motor command transition, suppose the obtained motor commands are forwarded to the actuator module in the form of PWM signals, with a 50 MHz system clock, the time spent on PWM signal transmission will be below 1 µs, without considering the path time. The power consumed by PWM generation is observed by monitoring the change in power consumption when the development board is in the idle state and in PWM transmitting state.

\Cref{tbl:benchmarkingColibriUAV} lists the time and power consumed by {ColibriUAV}. Altogether, 163 ms are spent from the event data readout to the PWM signals arriving at the flight control unit. Of this, a significant amount of time (\SI{131}{\milli\second}) is spent on spike preprocessing in the cluster. The latency here can be further decreased by memory usage optimization. The time spent on SNE is \SI{32}{\milli\second} for a complete SNN inference. The power consumed on the platform is \SI{46.98}{\milli\watt}, including the camera, the Kraken SoC, and the interfaces, and only \SI{9.2}{\milli\joule} for each complete loop. As far as we know, this work is the first to describe the latency and power performance of a complete, closed-loop neuromorphic UAV platform that can handle event data from a DVS camera onboard.

\begin{table}[!t]
\caption{Performance of ColibriUAV on reference neural network}
\label{tbl:benchmarkingColibriUAV}
\centering
\begin{threeparttable}
\begin{tabularx}{\linewidth}{{ p{2.3cm} p{1.7cm} p{1.7cm} p{1.8cm}}}
\hline
Modules\tnote{a}  & Latency  (\si{\milli\second}) &Power (\si{\milli\watt})\tnote{b} & Energy (\si{\milli\joule})\tnote{b} \\\hline
DVS and SAER (single event-frame) & 0.069 &  11.076\tnote{c} & 0.0008 \\ \hline
DVS and SAER (one window / 4350 event frames) & 300 &  11.076 & 3.323 \\ \hline
Preprocessing (Cluster) & 131 & 34 & 4.5 \\\hline
Inference (SNE) & 32 & 44 & 1.4 \\\hline
PWM(50\% duty) & 	$<$0.001 & 0.3 & 	$<$0.001 \\\hline
\textbf{Total} & 163 \tnote{d}  & 46.98\tnote{e} & 9.224 \\\hline

\end{tabularx}
\begin{tablenotes}
    \item[a] Network is deployed on SNE with input events window size of \SI{300}{\milli\second}. 
    \item[b] Power and energy are observed in the active status of the system dealing with one event frame in a pipeline way, namely the whole power consumed on all components.
    \item[c] Where is camera itself consumes 0.420 \si{\milli\watt}.
    \item[d] Data sensing and transmission is in parallel with SNE inference, thus not accumulated.
    \item[b] Power and energy are observed in the active status of the system dealing with one event frame in a pipeline way, namely the whole power consumed on all components.
    \item[e] The average total power consumption during inference is 35.6 \si{\milli\watt}.
\end{tablenotes}
\end{threeparttable}
\end{table}

\section{Conclusion}
This work describes ColibriUAV, a milliwatts drone platform that uses the Kraken SoC with an embedded neuromorphic accelerator and the benchmarks for the latency and power consumption of various components in the event signal processing pipeline and concludes that the closed-loop neuromorphic approach from event perception to decision-making offers promising performance for drone applications. The paper also presents a state-of-the-art event data fetching interface that is much faster and uses significantly less power than the USB approach used in existing DVS-equipped UAVs. In future work, the authors plan to investigate complete onboard event-based autonomous navigation, including real-time target tracking, obstacle recognition and avoidance, and optical flow estimation, using the proposed UAV platform.

\vspace{-0.5mm}



\bibliographystyle{IEEEtran}
\bibliography{sample}

\end{document}